\documentclass[letterpaper, 10 pt, conference]{ieeeconf}  % Comment this line out if you need a4paper

\IEEEoverridecommandlockouts                              % This command is only needed if 
\usepackage{amsmath,amssymb,amsfonts,amscd}
\usepackage{float}
\usepackage{graphicx}
\usepackage{epstopdf}
\usepackage{color}
\usepackage{verbatim}
\usepackage{tikz,times}
\usepackage{xcolor}
\usepackage{lipsum}
\usepackage{makecell}
\usepackage{booktabs}
\usepackage{algpseudocode}
\usepackage{tabularx}
\usepackage{cellspace}
\usepackage[Export]{adjustbox}
\usepackage{caption}
\usepackage{cuted}
\usepackage{pifont}% http://ctan.org/pkg/pifont
\usepackage{bbm}
\usepackage{gensymb}
\usepackage[ruled,vlined]{algorithm2e}
\usepackage{subcaption}

\makeatletter
\let\NAT@parse\undefined
\makeatother
\usepackage{lipsum}
\usepackage{hyperref}
\hypersetup{
  colorlinks = true,
  linkcolor = red,
  anchorcolor = red,
  citecolor = green,
  urlcolor = blue,
  pdftitle={WATonoBus Autonomous Shuttle: An Overview of System Hardware and Software Architecture and Functionalities}
}
\usepackage{stfloats}
\usepackage[capitalize]{cleveref}
\newcommand{\orcid}{\hspace{0.5mm}\includegraphics[height=\fontcharht\font`A]{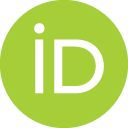}}

\title{\LARGE \bf
WATonoBus: Field-Tested All-Weather Autonomous Shuttle Technology
}

\author{Neel P. Bhatt$^{\href{https://orcid.org/0000-0002-1807-9672}{\orcid}}$,
        Ruihe Zhang$^{\href{https://orcid.org/0000-0002-2902-807X}{\orcid}}$,
        Minghao Ning$^{\href{https://orcid.org/0000-0003-4333-5524}{\orcid}}$,
        Ahmad R. Alghooneh$^{\href{https://orcid.org/0009-0009-6983-2511}{\orcid}}$,
        Chen Sun$^{\href{https://orcid.org/0000-0001-8772-9627}{\orcid}}$,
        Pouya Panahandeh$^{\href{https://orcid.org/0009-0007-2459-0382}{\orcid}}$,\\
        Ehsan Mohammadbagher$^{\href{0009-0001-5908-8743}{\orcid}}$,
        Ted Ecclestone$^{\href{https://orcid.org/0009-0008-4552-4410}{\orcid}}$,
        Ben MacCallum$^{\href{https://orcid.org/0000-0003-4241-7915}{\orcid}}$,
        Ehsan Hashemi$^{\href{https://orcid.org/0000-0002-6236-7516}{\orcid}}$, 
        and Amir Khajepour$^{\href{https://orcid.org/0000-0002-1998-6100}{\orcid}}$% <-this % stops a space
\thanks{Neel P. Bhatt, Ruihe Zhang, Minghao Ning, Ahmad Reza Alghooneh, Chen Sun, Pouya Panahandeh, Ehsan Mohammadbagher, Ted Ecclestone, Ben MacCallum and Amir Khajepour are with the Mechanical and Mechatronics Eng. Department, University of Waterloo, 200 University Ave W, Waterloo, ON N2L3G1, Canada. Ehsan Hashemi is with the Mechanical Eng. Department, University of Alberta, 9211-116 Street NW, Edmonton, AB T6G1H9, Canada. (e-mail:\{npbhatt, r422zhang, minghao.ning, aralghooneh, chen.sun, pouya.panahandeh, ehsan.mbagher, ejecclestone, ben.maccallum, a.khajepour\}@uwaterloo.ca, ehashemi@ualberta.ca).}%
}

\begin{document}

\maketitle
\thispagestyle{empty}
\pagestyle{empty}

%%%%%%%%%%%%%%%%%%%%%%%%%%%%%%%%%%%%%%%%%%%%%%%%%%%%%%%%%%%%%%%%%%%%%%%%%%%%%%%%
\begin{abstract}

All-weather autonomous vehicle operation poses significant challenges, encompassing modules from perception and decision-making to path planning and control. The complexity arises from the need to address adverse weather conditions such as rain, snow, and fog across the autonomy stack. Conventional model-based single-module approaches often lack holistic integration with upstream or downstream tasks. We tackle this problem by proposing a multi-module and modular system architecture with considerations for adverse weather across the perception level, through features such as snow covered curb detection, to decision-making and safety monitoring. Through daily weekday service on the WATonoBus platform for almost two years, we demonstrate that our proposed approach is capable of addressing adverse weather conditions and provide valuable insights from edge cases observed during operation.

\end{abstract}

%%%%%%%%%%%%%%%%%%%%%%%%%%%%%%%%%%%%%%%%%%%%%%%%%%%%%%%%%%%%%%%%%%%%%%%%%%%%%%%%

%%%%%%%%%%%%%%%%%%%%%%%%%%%%%%%%%%%%%%%%%%%%%%%%%%%%%%
\section{Introduction} \label{sec:Intro}
%%%%%%%%%%%%%%%%%%%%%%%%%%%%%%%%%%%%%%%%%%%%%%%%%%%%%%

Autonomous shuttle bus technology has been steadily advancing in recent years \cite{bucchiarone2020autonomous}. The introduction of self-driving shuttle buses that can operate along predefined routes without the need for human drivers will enable a new generation of last-mile transportation solutions. Autonomous shuttle buses have the potential to reduce traffic congestion and improve traffic efficiency. By eliminating the need for drivers to operate the vehicle, human errors can minimized and road safety can be enhanced due to the corresponding reduction in collision frequency \cite{sun2023_traffic}. Through collaborations among universities, government bodies, and industry partners, numerous cities have initiated pilot programs to test and deploy autonomous shuttle buses in controlled environments \cite{lowspeedautoshuttle, sun2021_dimensionless, hilgarter2020public}. Of these, most are simply exploratory and are motivated by short-term objectives such as the collection of autonomous driving data, the evaluation of economic benefits, and the opportunity to increase public exposure to autonomous driving technology. Although autonomous shuttle buses provide many benefits in comparison to traditional transportation methods, the task of matching or exceeding the performance of a human driver under dynamic weather and lighting conditions remains a challenge. Related safety standards have been drafted to provide qualitative guidelines for testing the performance of autonomous shuttle bus projects \cite{iso_lsad}, however, autonomous shuttle bus projects in adverse weather are still rarely found, posing challenges on the practicability of this technology in such scenarios.

\subsection{Related Works}
In December 1997, Schiphol Airport in Netherlands unveiled one of the early instances of an "automated people mover" system, known as the ParkShuttle \cite{ParkShuttle}. This pivotal development laid foundation for testing autonomous driving capabilities within a public environment. Currently, numerous global companies have successfully manufactured pilot self-driving shuttle buses for commercial purposes. These firms include Apollo, EasyMile, Navya, and Olli \cite{iclodean2020autonomous, azad2019fully, goldbach2022towards}. The WEpod project was initiated in the Netherlands with the aim of enhancing understanding of autonomous vehicles (AVs) \cite{van2017automated} where two driverless shuttles navigated through the Wageningen University campus, interacting with traffic while maintaining a top speed of 25 km/hr. Mcity at University of Michigan was introduced in 2018 as the first driverless shuttle testing field in the United States. One goal of the Mcity project was to analyze how passengers, pedestrians, bicyclists, and other drivers interact with the shuttle and gauge consumer acceptance of the technology \cite{mcitydriverlessshuttle}. Another autonomous bus project in Oslo was introduced called ``smarter transport in the Oslo region (STOR)'' \cite{MOURATIDIS2021321, roche2019public}. However, the buses were not fully automated and could not distinguish between different objects. At Aalto University, a driverless shuttle project operated among other traffic participants for 29 days and 365 kilometers and with 522 total passengers \cite{salonen2019towards}. Nevertheless, the buses were sensitive to environment dynamics. Snowflakes, heavy rain, and even flying leaves caused frequent emergency stops. In a comparable endeavor in Finland, two shuttles covered a 1.5 kilometer path in a park with five stops, interacting with pedestrians and cyclists \cite{feys2020experience}. The Taiwanese government also examined the potential of autonomous transportation by employing 9-meter driverless buses. These buses were deployed on a designated 2.9 km route in Taichung where they successfully tackled tasks such as navigating an open intersection and executing multiple U-turns \cite{nesheli2021driverless}. UNICARagil was another university-driven project supported by specialists from different enterprises \cite{Eckstein:835327, 9748024, 9625720, 9345480, 9827081}. They created an entirely new vehicle from the ground up and incorporated various enhancements however did not address adverse driving scenarios.

To this end, we aim to make the following contributions:
\begin{itemize}
    \item A modular software architecture for an autonomous shuttle bus, first of its kind in Canada \cite{mathworksDevelopingADAS}, considering adverse weather conditions (rain, snow, fog) which is experimentally validated in real-world scenarios.
    \item A perception module with multi-modal sensor fusion for accurate object and snow-covered curb detection in adverse weather.
    \item A dependable localization module with GNSS-denied capabilities in challenging weather conditions.
    \item A decision-making module, supported by a dedicated safety module, incorporating an intelligent bus stop pullover/merging function specifically tailored for shuttle bus service.
\end{itemize}

The remainder of this paper is structured as follows. In Section II, the WATonoBus sensor suite is introduced. Section III then presents the WATonoBus software system, followed by several case studies and discussion on insights in Section IV. Finally, Section V concludes the paper. 

%%%%%%%%%%%%%%%%%%%%%%%%%%%%%%%%%%%%%%%%%%%%%%%%%%%%%%
\section{WATonoBus Sensors and Calibration} \label{sec:WATonoBus Sensor and Calibration}
\begin{figure}[t]
\centering
\includegraphics[width=0.5\columnwidth]{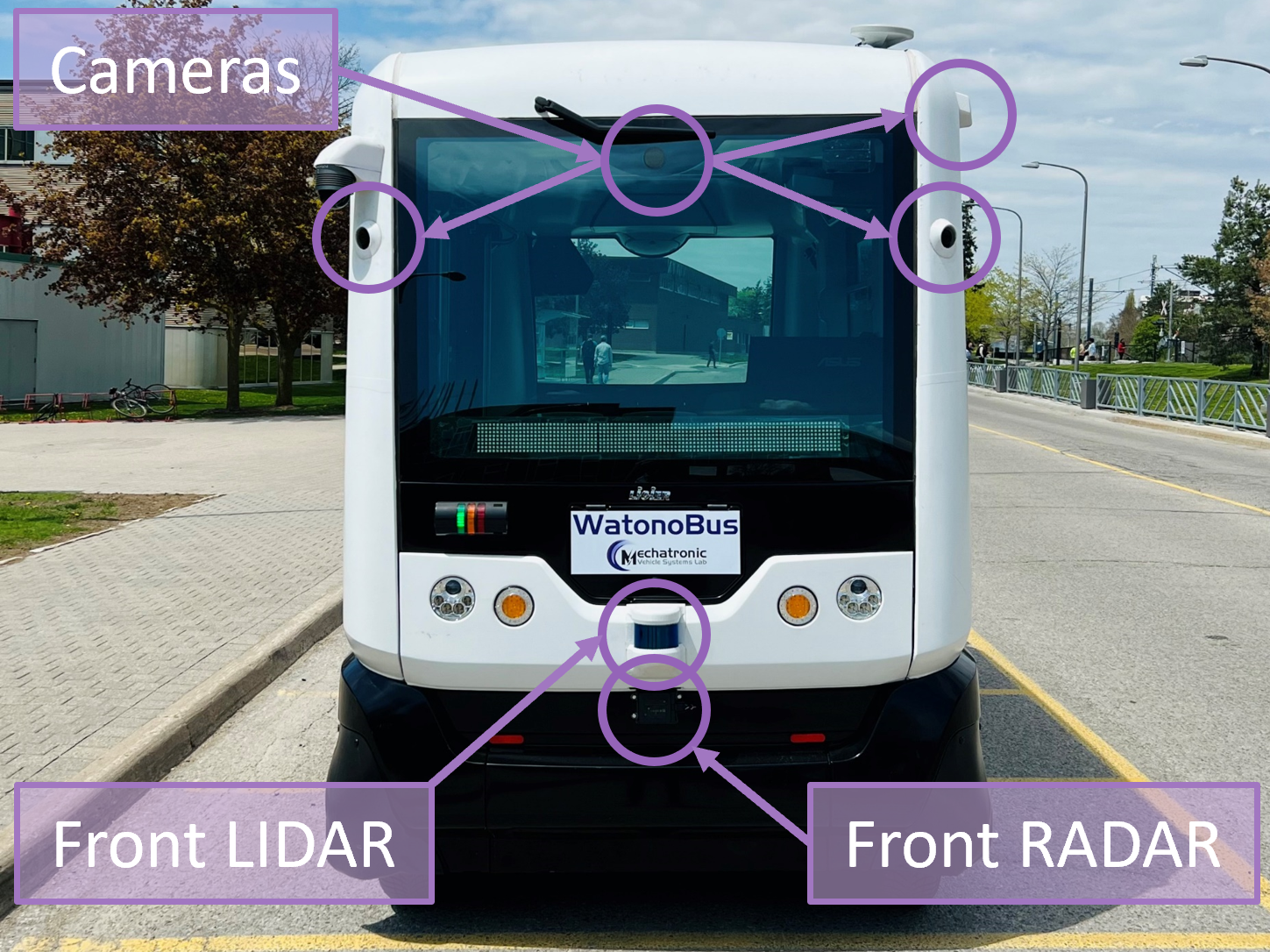}%
\includegraphics[width=0.5\columnwidth]{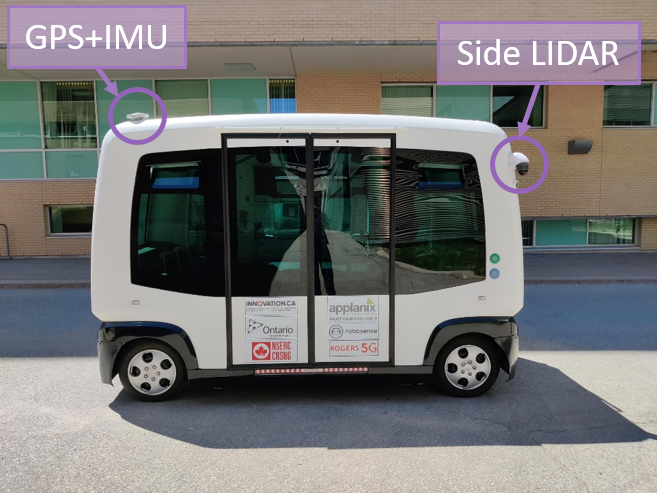}%
\hspace{0.5mm}%
\includegraphics[width=1\columnwidth]{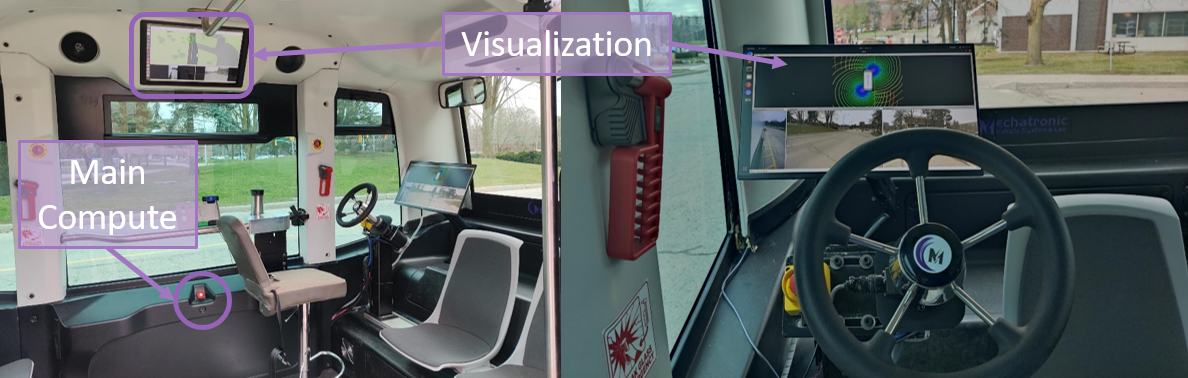}%
\caption{Illustration of WATonoBus sensor suite, compute system, control interface, and visualization utilities.} \label{fig:WATonoBus_sensors}
\end{figure}

%%%%%%%%%%%%%%%%%%%%%%%%%%%%%%%%%%%%%%%%%%%%%%%%%%%%%%
WATonoBus (\href{https://youtu.be/L-2PIEYpzuQ}{\textbf{video}}) is equipped with sensors placed in a configuration that allows sensing all surrounding objects as shown in Fig. \ref{fig:WATonoBus_sensors}.
%----------------------------------------------------------%
% \subsection{Sensors}
WATonoBus is instrumented with 6 3.2MP Basler Dart cameras, three 32 channel Robosense LiDARs, an Applanix POS LVX GNSS, and two Continental ARS408-21 radars to adequately detect any obstacle in its vicinity.
%% WATonoBus Sensors
%----------------------------------------------------------%
%----------------------------------------------------------%
Moreover, the compute system consists of a NVIDIA Jetson Orin AGX embedded unit along with visualization and utility modules on an Intel NUC Ruby with information flow via the Robot Operating System (ROS) \cite{ros}.

% %----------------------------------------------------------%
% \subsection{Sensor Calibration}

% To perform LiDAR-camera calibration, we establish a six-degree-of-freedom extrinsic transform. Initially, LiDAR points are projected onto the image plane with a preliminary guess. Refinement is achieved through manual or algorithmic adjustments using projection error, identifiable qualitatively in the image or quantitatively via checkerboard corner positions. This method eliminates the need for specialized calibration boards. Calibration was successfully performed using a tool we developed  available here \url{https://github.com/Neel1302/LiDAR-camera-calibration}.

%% System Block Diagram
%-----------------------------------------------------%
\begin{figure*}[t]
\centering
\includegraphics[width=0.9\linewidth]{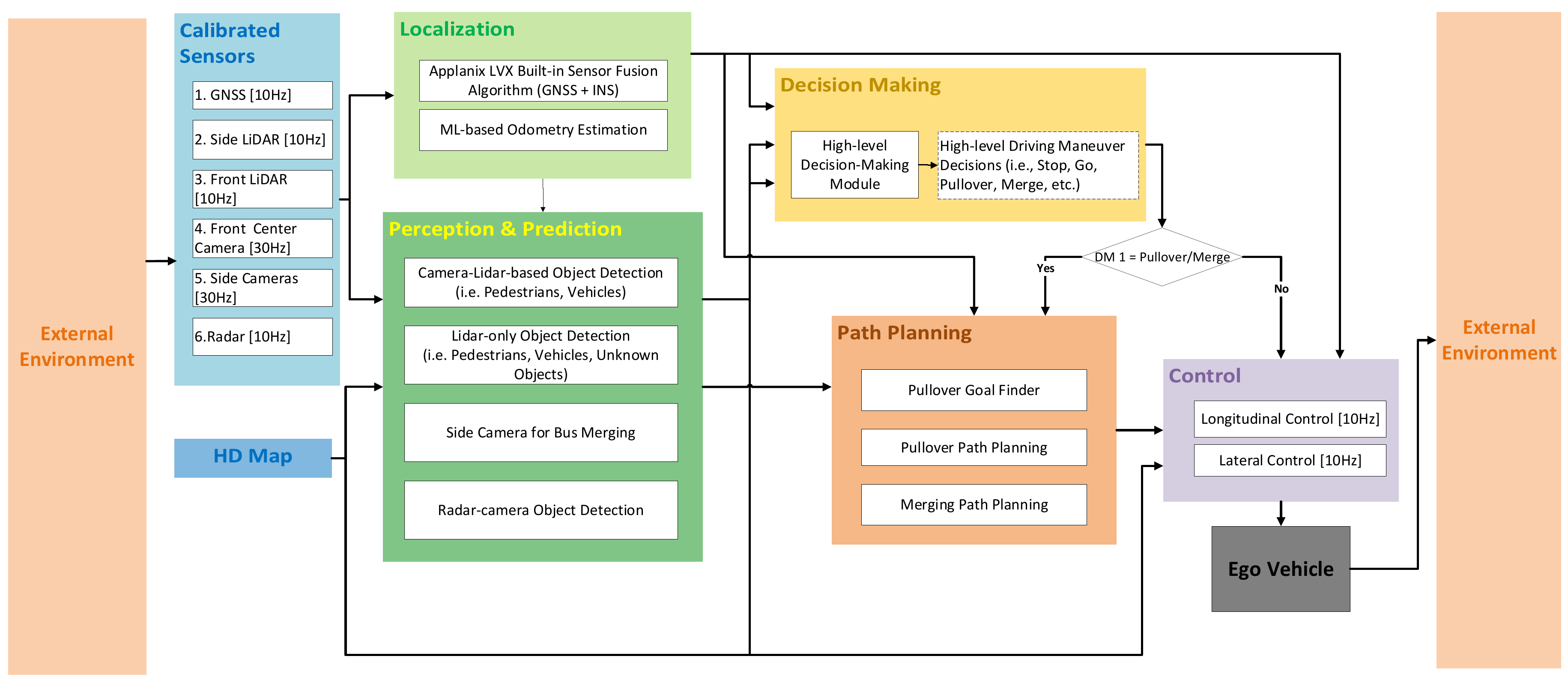}
\caption{A schematic of the overall software structure in WATonoBus.}
\label{fig:software_system_overview}
\end{figure*}
%-----------------------------------------------------%
% To calibrate multiple LiDARs, we use the Iterative Closest Point (ICP) algorithm \cite{zhang1994iterative}.

% Compensating for ego vehicle motion in LiDAR data processing is crucial to address distortions introduced by the vehicle's movement during the LiDAR sampling period. To mitigate this, a motion compensation method is proposed, leveraging high-frequency Inertial Navigation System (INS) data. Motion compensation involves establishing the transformation between the recorded pose and the pose at the end of the sampling period using INS data. This transformation, applied within 5ms, corrects for ego motion, enhancing the accuracy of the point cloud representation and improving subsequent data analysis and processing tasks.

%%%%%%%%%%%%%%%%%%%%%%%%%%%%%%%%%%%%%%%%%%%%%%%%%%%%%%
\section{WATonoBus Software System} \label{sec:WATonoBus_Software_System}
%%%%%%%%%%%%%%%%%%%%%%%%%%%%%%%%%%%%%%%%%%%%%%%%%%%%%%

In this section, we propose a multi-modal and modular software architecture with considerations for adverse weather conditions. A high-level system diagram is shown in Fig. \ref{fig:software_system_overview}.
%----------------------------------------------------------%
\subsection{Perception} 
%-----------------------------%
\subsubsection{2D Camera Detection}
You only look once (YOLO) v4 stands out as a real-time object detector, showcasing a notable advancement in the YOLO detector series by achieving enhanced accuracy and speed compared to its predecessors \cite{Bochkovskiy2020YOLOv4OS}. The architecture of YOLOv4 comprises of three key components: the backbone (CSPDarknet53), the neck (PANet and SAM block), and the head (YOLOv3 head). These components efficiently extract and fuse features, contributing to the model's performance. Furthermore, YOLOv4 employs various state-of-the-art techniques such as Mish activation, CIOU loss, and DropBlock regularization, alongside features such as multiple anchor boxes per grid cell and three scales for detection. Recognizing these attributes, we developed a YOVOv4 ROS package and retrained the network with domain-specific data for our campus environment (\href{https://tinyurl.com/yolo-ros-1}{\textbf{code}}).
%-----------------------------%
\subsubsection{LiDAR-only Detection}
\begin{figure}[t]
\centering
  \includegraphics[width=0.8\linewidth]{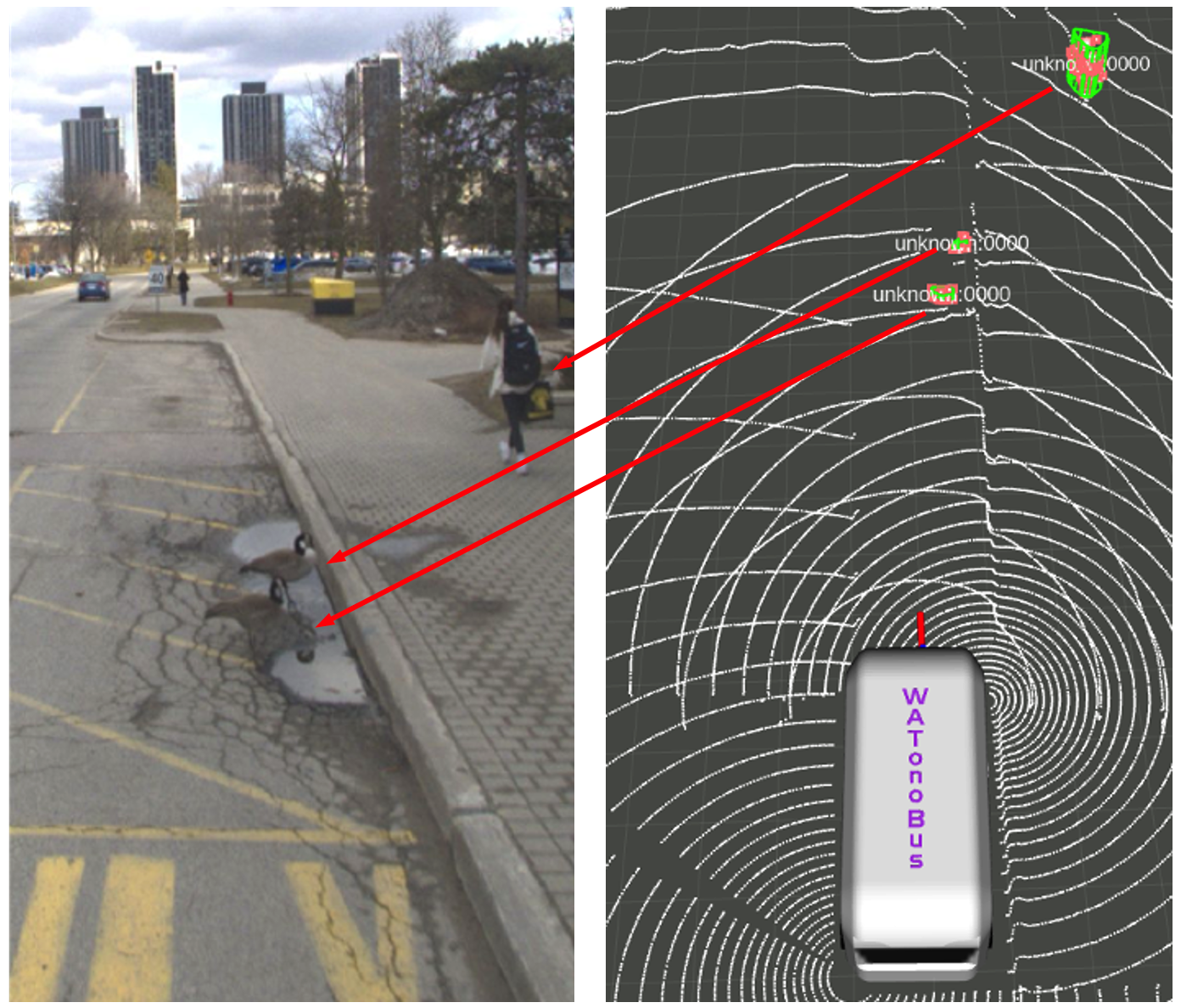}
\caption{An example for geese detection. Points from the geese are correctly segmented as shown in red, and their corresponding convex hulls are shown in green polylines.}
\label{fig:lidar_only_goose_detection}
\end{figure}

\begin{figure}[t]
  \includegraphics[width=\linewidth]{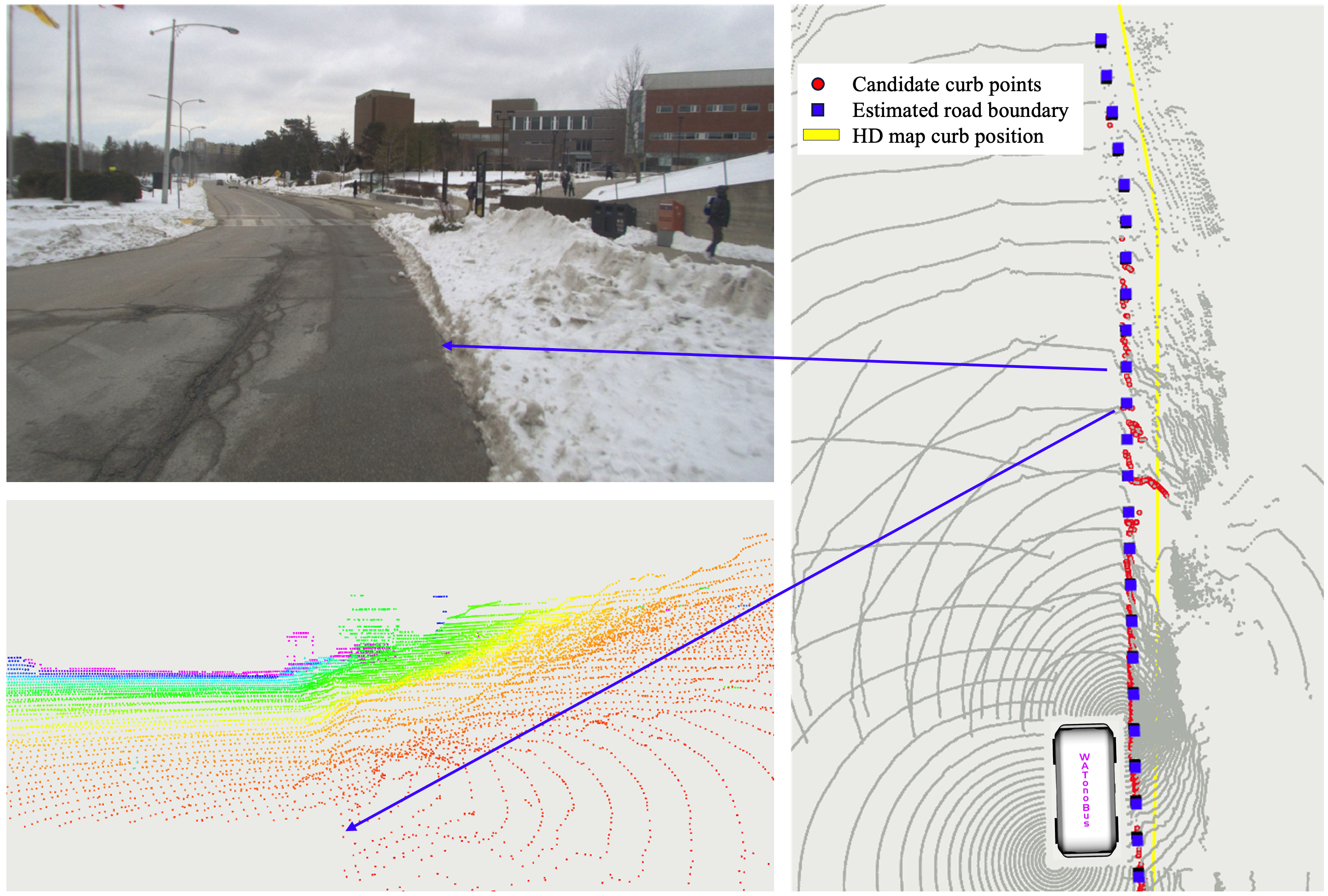}
\caption{(Upper left) Camera view. (Lower left) Point cloud view.  Red points denote closer distance to ego. (Right) Road boundary detection results in heavy snow. Red points are detected curb points from LiDAR point cloud. Blue squares are estimated road boundary. Yellow line denotes curb position from HD map. The blue boxes are on the left of the yellow line, reflecting actual drivable space.}
\label{fig:adaptive_curb_detection}
\end{figure}

LiDAR-only detection provides a reliable and accurate enhancement to camera-based detection, particularly for uncommon object classes and curb detection under adverse weather conditions. It begins with the point cloud concatenation of the front and two side blind-spot LiDARs. Firstly, an adaptive grid ground segmentation algorithm is employed to remove ground points. Leveraging high definition (HD) map information, we divide the combined point cloud into multiple regions,  allowing for a more accurate and robust estimation of road surfaces. This is achieved by adjusting road surface fitting thresholds based on distance from the ego vehicle, employing stricter thresholds for closer areas and more lenient thresholds for regions further away. This approach effectively enhances the model's robustness against noise attributed to ego vehicle motion or erroneous point cloud reflections, ensuring the detection of all points above the road surface for subsequent object detection. Post ground segmentation, an adaptive noise point detection algorithm filters out raindrops or snowflakes (\href{https://youtu.be/avKjMzzz0XQ}{\textbf{video 1}} and \href{https://youtu.be/FePtm6bRhWA}{\textbf{2}}). Density-based spatial clustering of applications with noise (DBSCAN) algorithm is employed for object clustering \cite{10.5555/3001460.3001507}, enhancing LiDAR-based detection's robustness in identifying unfamiliar objects. An example for geese detection is shown in Fig. \ref{fig:lidar_only_goose_detection} (\href{https://youtu.be/khnOvfHGWDw}{\textbf{video 1}} and \href{https://youtu.be/bJooilAlqys}{\textbf{2}}). The algorithm identifies obstacle points, selects candidate points close to the ground, and incorporates additional data points in the absence of LiDAR information. For varying curb shapes, a lateral residual model is used. For safety, candidate points on the left of the fitted curb model are reassigned as obstacles, aiding in detecting objects like geese near the curb.
In addition to uncommon object detection, LiDAR-only detection can perceive snow-covered curbs by employing HD maps to partition potential curb points in the point cloud into adjacent road regions. This segmentation provides accurate driving boundary constraints despite a mismatch in the HD map information due to positioning errors or due to snow. In scenarios where snow covers the curbs and leads to tighter road boundaries, a weight factor is introduced during road surface model fitting to mitigate errors. Fig. \ref{fig:adaptive_curb_detection} exemplifies our approach's effectiveness in curb detection in heavy snow, showcasing a comparison between camera view and point cloud, revealing drivable space detection results \href{https://youtu.be/xNh1ojU7LgI}{(\textbf{video})}.
%-----------------------------%
\subsubsection{Radar-only Detection} % Ahmad
Radar-only detection offers another level of robustness in the perception module since it offers reliable operation under various weather conditions, especially in poor visibility conditions. In addition, the wide field of view and long-range capabilities (up to 250m) of the selected radar sensors enable anticipatory driving, allowing autonomous vehicles to adjust speed or change lanes in response to upcoming traffic. Radar also possesses inherent advantages to distinguish between stationary and moving objects.

In addition to tracking and prediction, the Radar Cross Section (RCS) of radar-detected objects can be useful in object classification. This information can be useful for classifying  different objects, thereby enabling the ego vehicle to make safer decisions.

\begin{figure*}[t]
\centering
\includegraphics[width=0.8\linewidth]{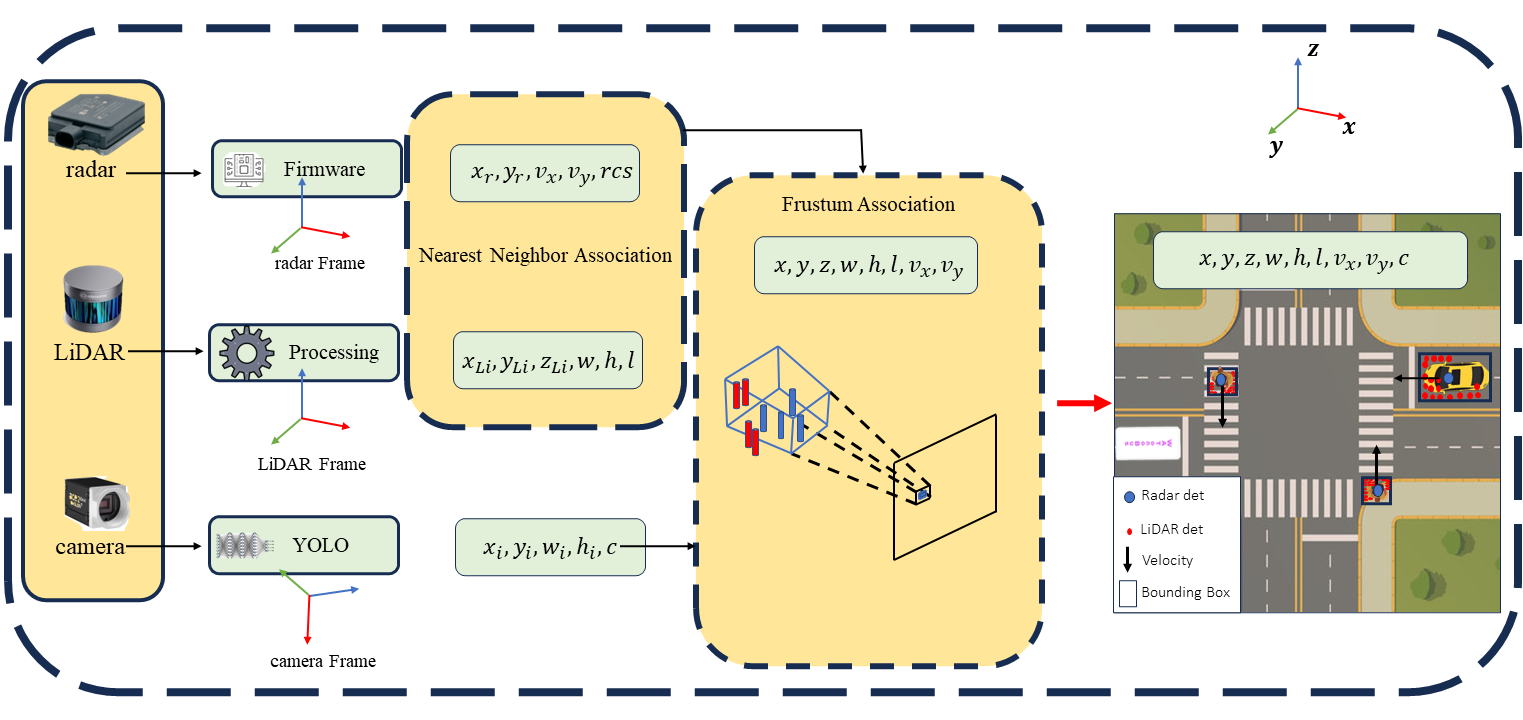}
\caption{A schematic of the overall late fusion algorithm. The radar provides longitudinal and lateral positions and velocities $x_r, y_r, v_x$, and $v_y$ of objects. After ground removal and clustering, we have object positions and the dimensions from LiDAR-only detection algorithm, $x_{Li}, y_{Li}, z_{Li}, w, h$, and $l$. From YOLOv4 and camera-based object detection we have bounding box dimensions and class $x_i, y_i, w_i, h_i$, $c$. The fusion process completes with frustum association and the final output is bounding box position, dimension, velocity and class $x,y,z,w,h,l,v_x,v_y$ for each object.}
\label{fig:fusion_block_diagram}
\end{figure*}

%-----------------------------%
\subsubsection{Fusion} % Ahmad
An object detection and tracking fusion algorithm is employed in WATonoBus to leverage the strengths of camera, LiDAR, and radar, combining their outputs for a more robust and accurate representation of the surrounding environment. Particularly, we employ a late fusion strategy, where each sensor's data is processed independently before being combined (See Fig. \ref{fig:fusion_block_diagram}). The first step involves associating the radar and LiDAR detection results. The radar-only detection algorithm estimates longitudinal and lateral velocities and RCS of detected objects, while the LiDAR-only detection algorithm generates a point cloud to indicate the shape and distance of these objects. After that, we project the combined radar and LiDAR information into the image space used by the camera-only detection algorithm. This aligns the object detection results with camera's depth priors, providing a coherent view of the objects in the environment in relation to the imagery captured. Following the frustum approach, the fusion is carried out based on the depth order. This ensures a more accurate and holistic representation of the detected objects in the 3D space, considering the depth information from the camera and the associated data from radar and LiDAR. The final output of the fusion algorithm is a comprehensive three-dimensional bounding box encapsulating each detected object, along with its estimated velocities. 

This late fusion approach allows us to fully leverage the strengths of each sensor, resulting in a more accurate and reliable object detection and tracking system. The radar fusion with LiDAR point cloud in the first step will enable the scheme to eliminate the false alarms both for radar and LiDAR in challenging weather conditions. Therefore, if the LiDAR-radar fused detections did not find a association with a camera bounding box, it will be labeled as an unknown object. This result combines the best of all sensor capabilities and offers a detailed understanding of each object's position, dimensions, and motion, enabling the shuttle to make well-informed navigation decisions, thereby significantly enhancing the safety and reliability of WATonoBus (\href{https://youtu.be/iQqLdEyVjdg}{\textbf{video}}). 

\subsection{Localization}
% A brief description of the "dependable" localization module:
WATonoBus requires precise positioning to safely navigate through the drivable space in different weather conditions. This involves accessing the vehicle’s current location and heading  while taking into account the location of road boundaries and nearby landmarks through HD maps. Accordingly, a dependable localization module is designed with both GNSS-based and GNSS-denied capabilities.
%-----------------------------%
\subsubsection{GNSS-based Capability}
WATonoBus is equipped with an Applanix POS LVX GNSS-Inertial system aided with Real-Time Kinematic (RTK). It provides the position with sub-decimal accuracy and precise heading information of the vehicle.
%-----------------------------%
\subsubsection{GNSS-denied Capability}
In case of low accuracy or GNSS outage due to adverse weather conditions, the GNSS-denied capability provides a temporary localization solution for the vehicle to perform a safe stop maneuver. It fuses machine-learning-based odometry estimations with LiDAR detections of georefrenced nearby HD map landmarks such as, light poles, building planes, and road curbs to estimate the location and heading of WATonoBus. The use of detections of large nearby landmarks that stand out from the environment ensures reliable localization when the road is not visible (e.g., covered by snow).
%-----------------------------%
\subsubsection{HD Map}
To form the HD map, the entire drivable space is discretized into a finite number of equally spaced intervals along an $s$-curve, which is a one-dimensional path coordination system that describes the vehicle's location with a single scalar. At each $s$-coordinate, a record of different information is stored in a database that forms the HD map. When WATonoBus is operating in GNSS-denied conditions, the dependable localization module will retrieve the information of reliable landmarks associated with the current s-coordinate from the HD map database.
% \begin{figure}[t]
% \centering
% \includegraphics[width=0.8\linewidth]{figures/localization/WatanoBus_localization.jpg}
% \caption{The HD map contains different layers of information including the relative location of nearby important landmarks (e.g. curb, lane, light poles, building planes, signs, etc.), road type, intersection type, driving condition, etc, stored at each s-coordinate.}
% \label{fig:hd_map}
% \end{figure}

%----------------------------------------------------------%
\subsection{Decision Making}
\begin{figure*}[t]
\centering
\includegraphics[width=0.8\linewidth]{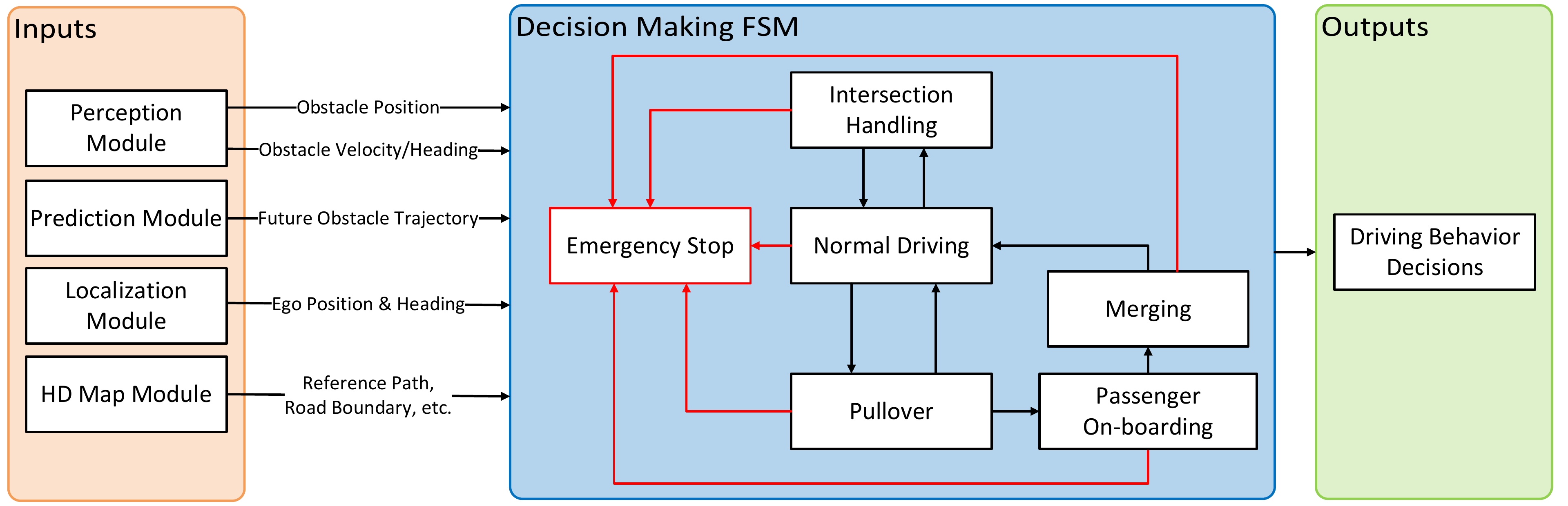}
\caption{FSM Block Diagram of WATonoBus DM Module.}
\label{fig:FSM_block_diagram}
\end{figure*}

The objective of the decision making module is to make safe driving decisions based on the information from perception, prediction, localization modules, and HD map \cite{dm_review, bhatt2023mpc, bhatt2023socially}. The high-level decisions will then serve as the inputs to the downstream modules to perform path planning, pullover/merging, and vehicle control. Within the service environment of WATonoBus, the driving behavior decisions are limited to Go/Stop decisions during normal driving and intersection handling, and additional Pullover/Merge, door open/close decisions at bus stop regions. Apart from normal driving decisions, the decision making module will also output emergency stop decisions as the safety fallback strategy. Furthermore, although all the interested obstacles will be reviewed through the decision making module, the final decision will be made with respect to the highest risk obstacle. To achieve the aforementioned objective, finite state machine (FSM) was chosen as the backbone in WATonoBus decision making module (See Fig.~\ref{fig:FSM_block_diagram}) because its strengths in structure modularity, interpretation transparency, and robustness to errors or unexpected events \cite{jaswanth2022autonomous}. 

The three important FSM components of the WATonoBus decision making module are states, state transition conditions, and decision-making logic in each state. In an FSM, a state represents a unique situation in the system. In this case, the states are defined based on the static road features in the bus service environment. Specifically, there are normal road segments, all-way stop T-shape intersections, and bus stop regions along the campus ring road. Therefore, the five major states in FSM are defined, namely normal driving, intersection handling, pullover, merging, and passenger on-boarding states. Moreover, to handle the unexpected events, an emergency state is added. A transition in the FSM is a change from one state to another. For WATonoBus decision making module, the transitions are mainly governed by the ego vehicle location and the perception results. Specifically, the ego vehicle location is used to compute the remaining distance to the coming static road features such as all-way stops or bus stop regions. For example, the FSM will change to the pullover state if the ego vehicle is approaching to a bus stop region and then a pullover path will be required. In addition, perception results contain the obstacles around the ego vehicle, indicating whether it is safe to resume bus normal driving. For instance, if there is no vehicle approaching from the rear, the FSM will switch from pullover state to merging state after waiting for passenger on-boarding. Lastly, decisions made under each state can base on different logic due to the traffic rules and vehicle control requirements. The traffic rules in each FSM state list the expected ego vehicle behaviors. To handle an intersection properly, a few traffic rules need to be satisfied in sequence. Assuming there is no pedestrian, the ego vehicle needs to first check if a stop is required for the intersection-waiting vehicle in its lane. It will then be required to stop at the intersection for a while till the intersection is clear. Depending on the arrival order, a decision of whether to yield for the waiting vehicle at other intersection entrances needs to be made. After clearing all these traffic rules check, the ego vehicle can finally start moving and let FSM switch from intersection handling to normal driving state. In addition to traffic rules, the decision-making logic in each state differ from the required vehicle control parameters. For instance, the maximum speed of merging is slower than that of normal driving to ensure the bus to carefully merge out of the bus region while be able to stop immediately for any vehicles suddenly come from the rear (\href{https://youtu.be/j8ZLZvTz2xE}{\textbf{video 1}} and \href{https://youtu.be/43JPrNccSvo}{\textbf{2}}).

%----------------------------------------------------------%
\subsection{Path Planning} %Pouya
WATonoBus relies on path planning to safely and efficiently navigate through the environment. This involves determining a path from the vehicle's current location to its intended destination while taking into account obstacles, traffic regulations, and other factors. There are two main components to path planning, namely global and local path planning.
%-----------------------------%
\subsubsection{Global Path Planning} %Pouya
The global planner determines a high-level path from the current vehicle location to the destination, utilizing HD map information. This algorithm ensures a safe and efficient path, considering any objects or environmental constraints. As WATonoBus operates on a one-lane road, the global planner defines the right curb and center line as the right and left boundaries of the drivable space. The center of this space is assigned as the reference path for the bus to track.
%-----------------------------%
\subsubsection{Local Path Planning} %Pouya
The task of local path planning is to come up with a more detailed path that follows the global path while taking into account the current state of the vehicle and any obstacles in its immediate surroundings. Moreover, this involves devising a path for the WATonoBus to safely execute pullover and merging maneuvers as needed. These are typically done using potential field \cite{Rasekhipour2017} and Bezier curves \cite{Choi2008}. With the potential field method, we have a mathematical framework to adjust the reference path of the bus whenever obstacles are detected within its drivable space. Following this, smooth segments are generated using Bezier curves from the modified waypoints obtained through the potential field method. For the pullover and merging maneuver, the waypoints are produced using the Bezier curves and control points. By combining the potential field with Bezier curves, the WATonoBus can quickly and safely navigate through its environment and execute its missions.

While following the local path, the objective of the motion controller is to select a desired speed and steering angle for the system to track at each sampling instant. This enables the vehicle to follow the reference path while safely interacting with obstacles (\href{https://youtu.be/7gQO7QoLIAA}{\textbf{video}}). The longitudinal controller utilizes a feed-forward algorithm that considers the position and velocity of the ego vehicle relative to obstacles. Firstly, the position $s$ of the WATonoBus and other obstacles along the path is estimated. The distance difference $\Delta s$ between the bus and obstacles along the path is then calculated. If there are no obstacles in the vicinity of the bus, the desired velocity selected by the controller is updated incrementally at each sampling instant using
\begin{equation}
    V_\text{des}(k) = V_\text{des}(k - 1) + a_\text{nom} \Delta t
\end{equation}
where $V_\text{des}(k)$ is the desired velocity at the sampling instant $k$, $a_\text{nom}$ is the nominal desired acceleration, and $\Delta t$ is the sampling time. This relationship enables the bus to freely accelerate or decelerate towards the reference velocity of 20 km/h. The sign of $a_\text{nom}$ depends on whether the vehicle's velocity is above or below 20 km/h, and $V_\text{des}$ is held at 20 km/h once the vehicle reaches this setpoint. In the presence of obstacles, the relationship remains similar; however, the nominal acceleration is replaced by a desired acceleration $a_\text{des}$ that the vehicle must achieve if it is to maintain a minimum spacing of $\Delta s_\text{min}$ from the obstacle. In this case, the desired velocity is updated as
\begin{equation}
     a_\text{des}(k) = \frac{-V_\text{des}(k - 1)^2}{2(\Delta s(k) - \Delta s_\text{min})},
 \end{equation}
 \begin{equation}
    V_\text{des}(k) = V_\text{des}(k - 1) + a_\text{des}(k) \Delta t.
\end{equation}

\begin{figure}[t]
\centering
\includegraphics[width=0.85\linewidth]{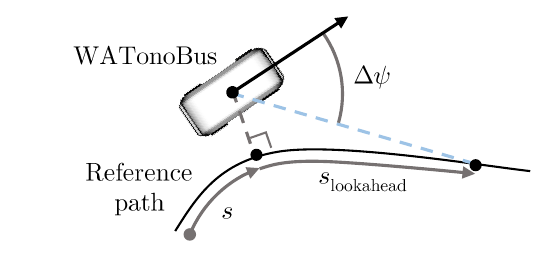}
\caption{Heading error definition.}
\label{fig:heading_error}
\end{figure}

The lateral controller considers the vehicle's path tracking error to select desired steering angles that enable the position of the vehicle to converge with the path \cite{samuel2016review, park2014development, ahn2021accurate}. After determining the position $s$ of the vehicle, a "look ahead" point along the path at the position $s_\text{lookahead}$ ahead of the bus is selected; the distance between $s$ and $s_\text{lookahead}$ is referred to as the look ahead distance, and it is selected by considering the velocity of the vehicle. 
Fig.~\ref{fig:heading_error} displays how the the vehicle's position and the position of the look ahead point are used to compute the vehicle's heading error with respect to the path, $\Delta \psi$.
The heading error $\Delta \psi$ then serves as the input to a PID controller, where the reference heading error $\Delta \psi_\text{ref}$ is 0; the output of the controller is the desired steering angle $\delta_\text{des}$. As the WATonoBus has all wheel steer functionality, $\delta_\text{des}$ serves as the desired angle for the front wheels while $-\delta_\text{des}$ is the desired angle for the rear wheels.

%----------------------------------------------------------%
\subsection{Safety Module}

%-----------------------------%
% \subsubsection{Safety Testing and Validation}
WATonoBus prioritizes system safety through offline validation, real-time monitoring, and continuous engineering with black-box recording for edge cases during deployments. Both software and hardware components adhere to Functional Safety and SOTIF guidelines \cite{iso_21448}. In addition, Failure Tree Analysis (FTA) and Systems-Theoretic Process Analysis (STPA) are used to assess potential failure cases and establish safety goals for the designated operational area. 

% To encompass various challenging cases, we developed a scenario generator to explore potentially dangerous movements of vehicles and pedestrians. It considers various configuration parameters like the number of traffic participants, their motion and velocity profiles. We explore different combinations of these configuration parameters in testing scenario generation and then examine the performance of decision making and control modules for any potential failures.

% %-----------------------------%
% \subsubsection{Safety Monitor and E-Stop}
% During real-time operation, system safety is maintained through reliability monitoring and fail-safe modules. Reliability monitoring module checks location, weather, and sensor performance, generating malfunction signals within the predefined Operational Design Domain (ODD) \cite{sun_2022odd}. These signals indicate the current estimation of system reliability in perception, localization, and planning. The fail-safe module responds to warnings by prompting the operator to take over or initiating an emergency stop based on warning priority.

%%%%%%%%%%%%%%%%%%%%%%%%%%%%%%%%%%%%%%%%%%%%%%%%%%%%%%
\section{Learning from Daily Operation} \label{sec:Learning_from_daily_operation}
%%%%%%%%%%%%%%%%%%%%%%%%%%%%%%%%%%%%%%%%%%%%%%%%%%%%%%
We present the handling procedure and lessons learned from both regular bus service and severe weather encountered in WATonoBus daily operations, followed by  discussing the insights gained from these case studies.

%----------------------------------------------------------%
\subsection{Bus Stop Handling}

\begin{figure}
  \begin{subfigure}{2cm}
    \centering\includegraphics[width=1\linewidth]{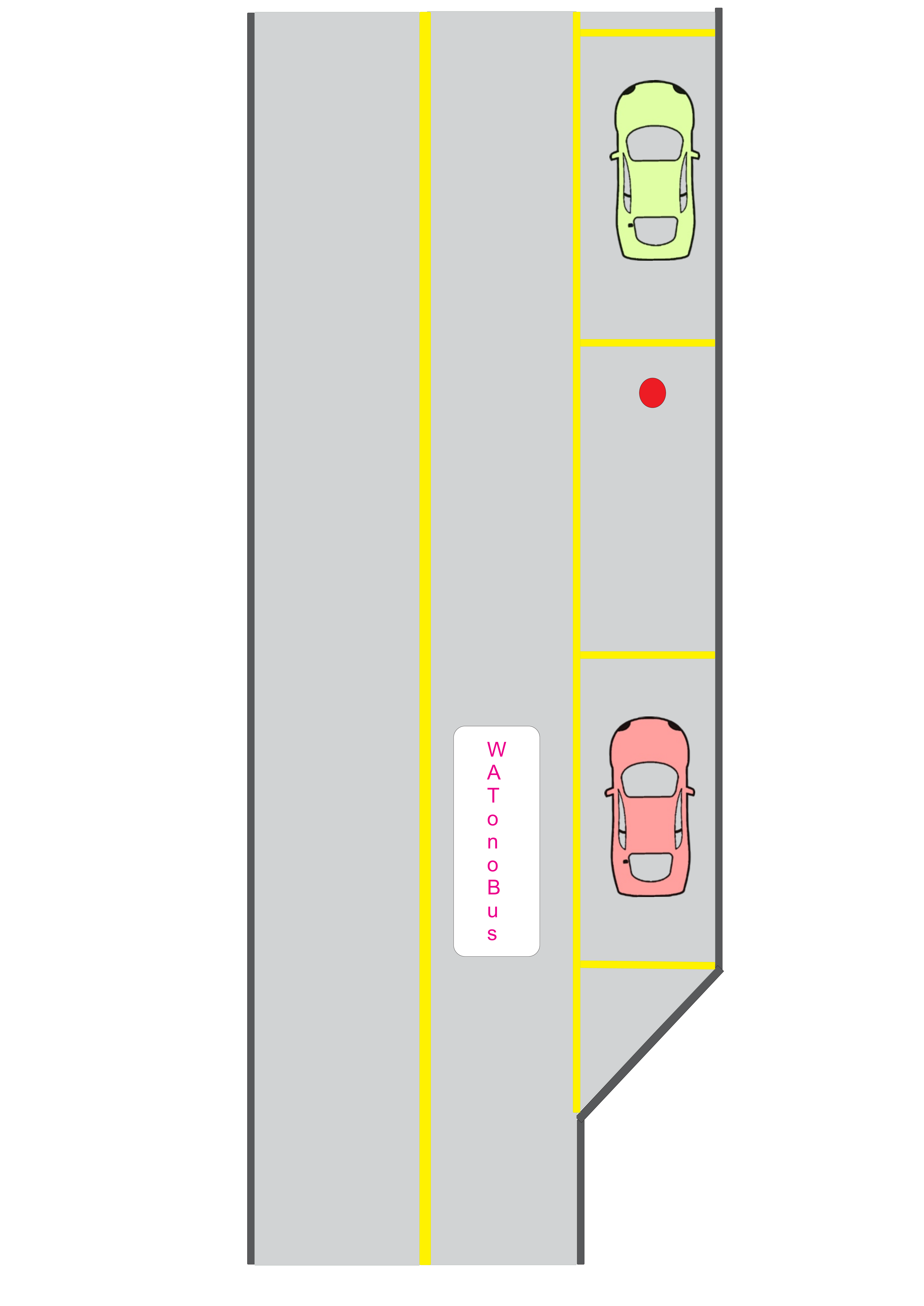}
    \caption{Selection}
  \end{subfigure}
  \begin{subfigure}{2cm}
    \centering\includegraphics[width=1\linewidth]{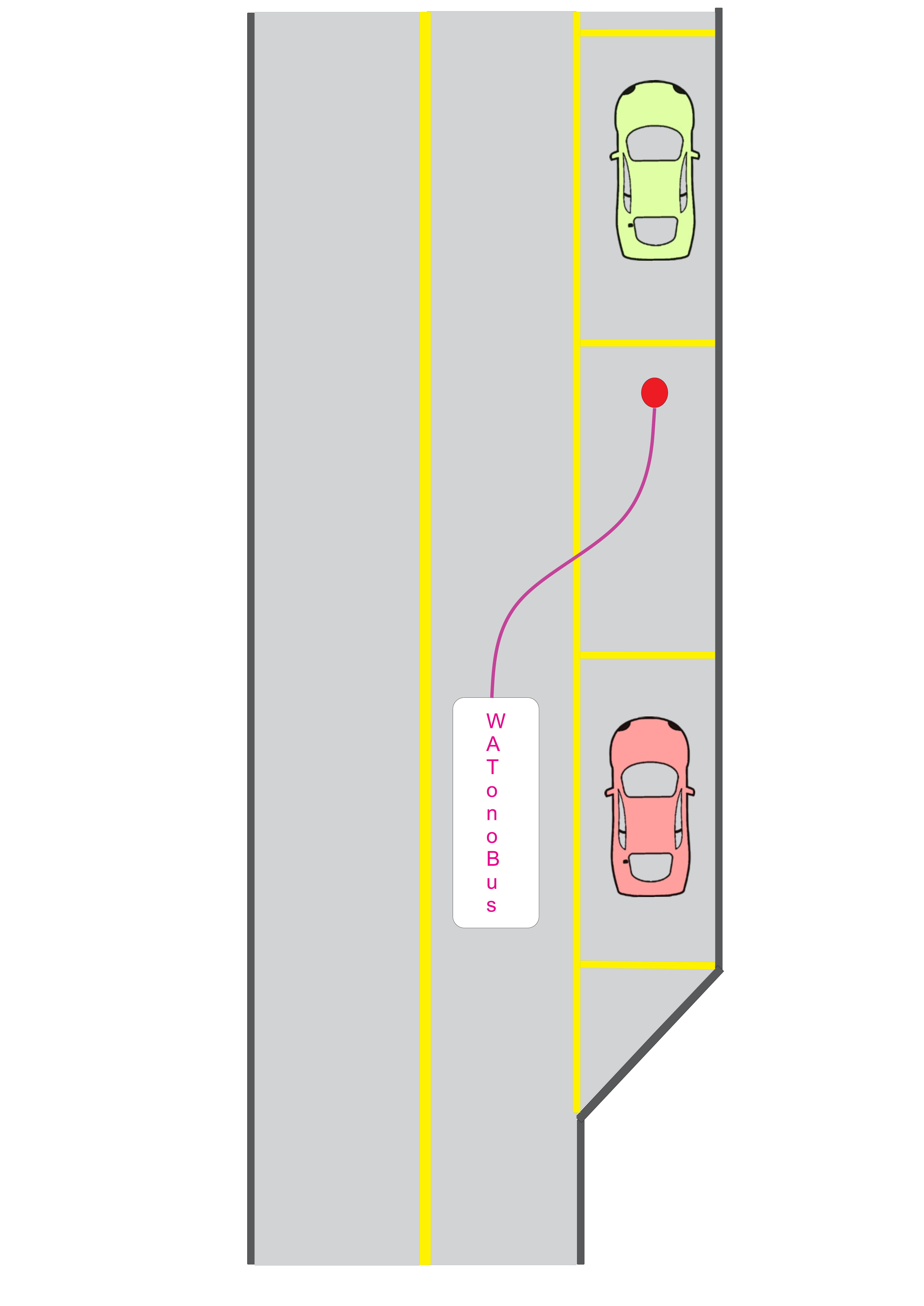}
    \caption{Planning}
  \end{subfigure}
  \begin{subfigure}{2cm}
    \centering\includegraphics[width=1\linewidth]{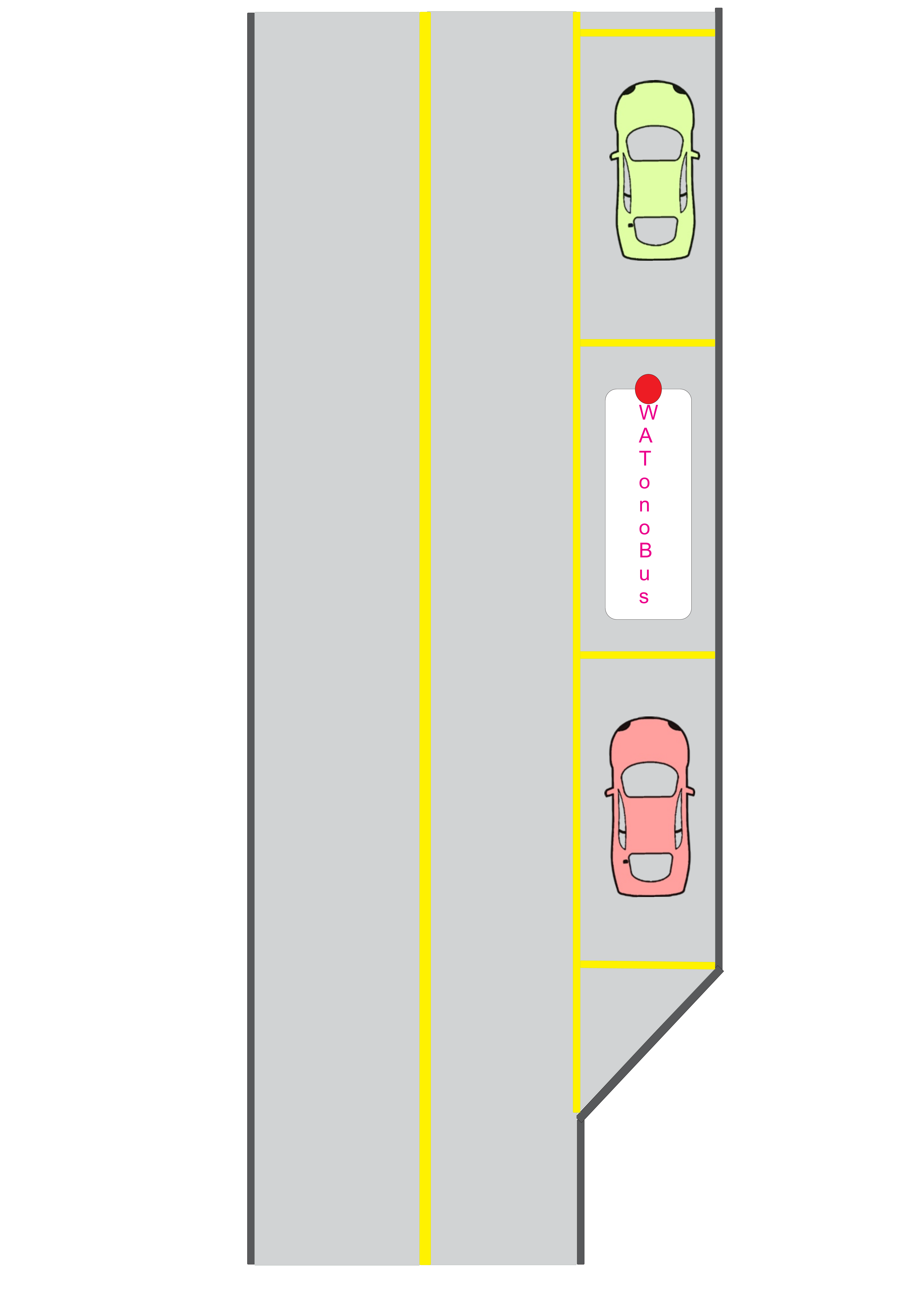}
    \caption{Boarding}
  \end{subfigure}
    \begin{subfigure}{2cm}
    \centering\includegraphics[width=1\linewidth]{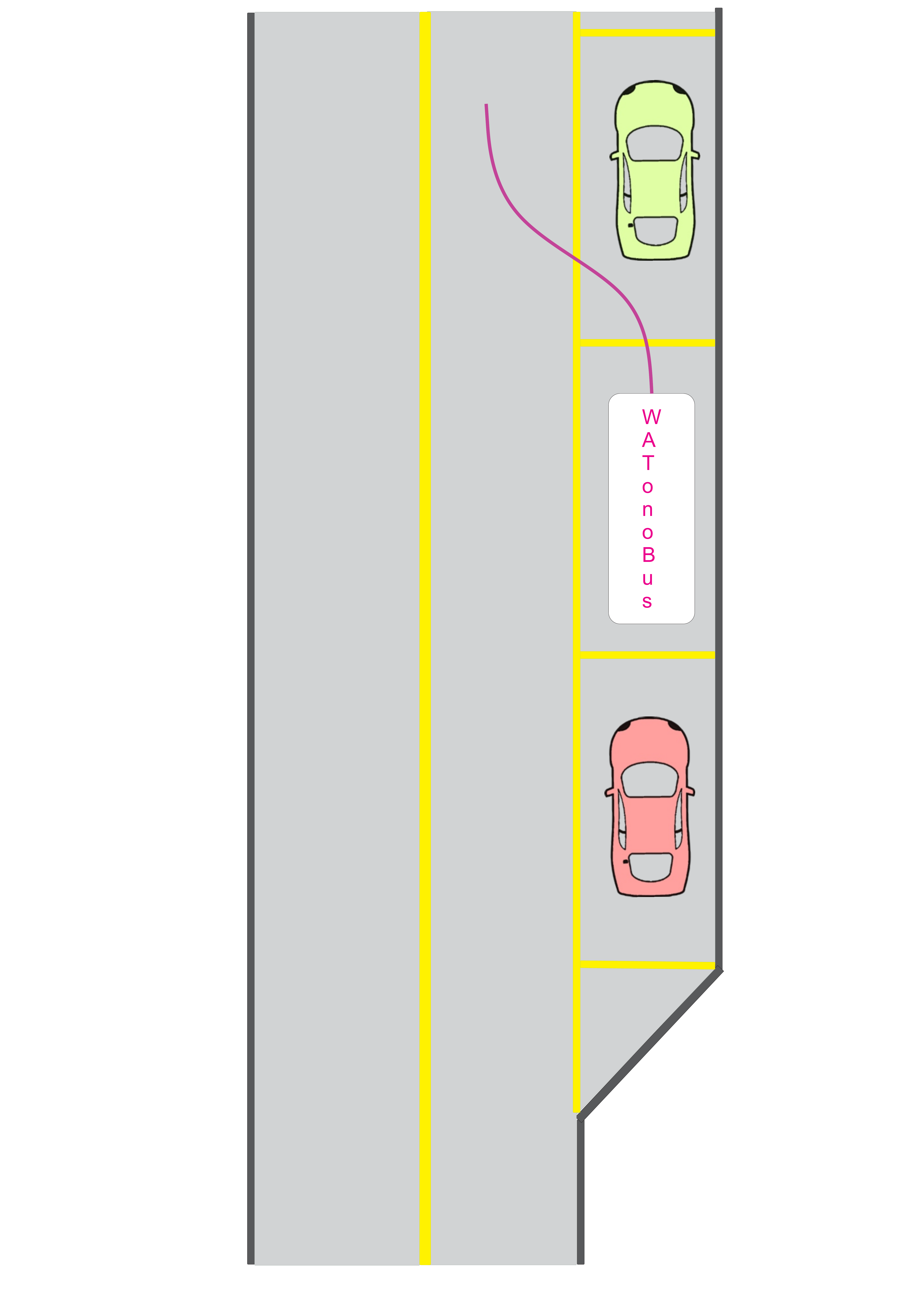}
    \caption{Merging}
  \end{subfigure}
  \caption{Pullover and merging procedure}
  \label{Figure: PulloverAnd Merging}
\end{figure}

The bus stop handling task involves four functional modules: dynamic spot identification, passenger on/off boarding, merging handling, and pullover/merging path planning modules (See Fig. \ref{Figure: PulloverAnd Merging}). The dynamic pullover spot identification module is crucial for ensuring safe and precise stopping at bus stops. Triggered by the decision-making (DM) module when the shuttle approaches a bus stop, this module conducts a careful scan of the area within a longitudinal distance of up to 20 meters from the shuttle's current position. It checks for any obstacles, such as stationary vehicles, pedestrians, or bollards, that may obstruct the shuttle's path during pullover. The module generates a specific point indicating a safe pullover spot, which is then relayed to the DM module and further to the path planner and control modules for navigation to the designated spot.

% The passenger on/off boarding module ensures a comfortable bus service experience. After pullover at the bus stop, the shuttle automatically opens the door and uses the in-door sensor to detect potential passengers attempting to board or alight. After a continuous check and waiting for a sufficient time, the shuttle locks the door and transitions to the merging status.

% The merging handling module combines rear radar and camera results for a comprehensive view of the traffic behind the ego vehicle, improving hazard detection and reaction capabilities. The radar provides reliable long-range detection, crucial for monitoring approaching traffic. It not only detects vehicles but also estimates their velocities, aiding in understanding their movement patterns. The rear camera complements these capabilities, serving as the final vote to prevent radar false alarms during merging. This fused data informs the vehicle's DM process, calculating the optimal timing and speed for a smooth merge into traffic. Anticipating potential conflicts by considering the speed and trajectory of approaching vehicles, our system adjusts its strategy for safety, ensuring the well-being of shuttle bus passengers and other road users.

The pullover/merging path planning module ensures a smooth path for the ego vehicle during pullover at bus stops or merging maneuvers (\href{https://youtu.be/FAVN4vSNxNQ}{\textbf{video}}). Specific waypoints are crucial during pullover, determined considering the bus's current position, heading, and the identified parking location. Bezier curves are utilized to create a smooth path between these waypoints. Similarly, for merging, waypoints are determined based on the bus's starting position, orientation, and the target location along its reference path. These waypoints guide the creation of a smooth path using Bezier curves. %Pouya

%-----------------------------%
\subsection{Intersection Handling}
\begin{figure}
  \begin{subfigure}{4cm}
    \centering\includegraphics[width=3.5cm]{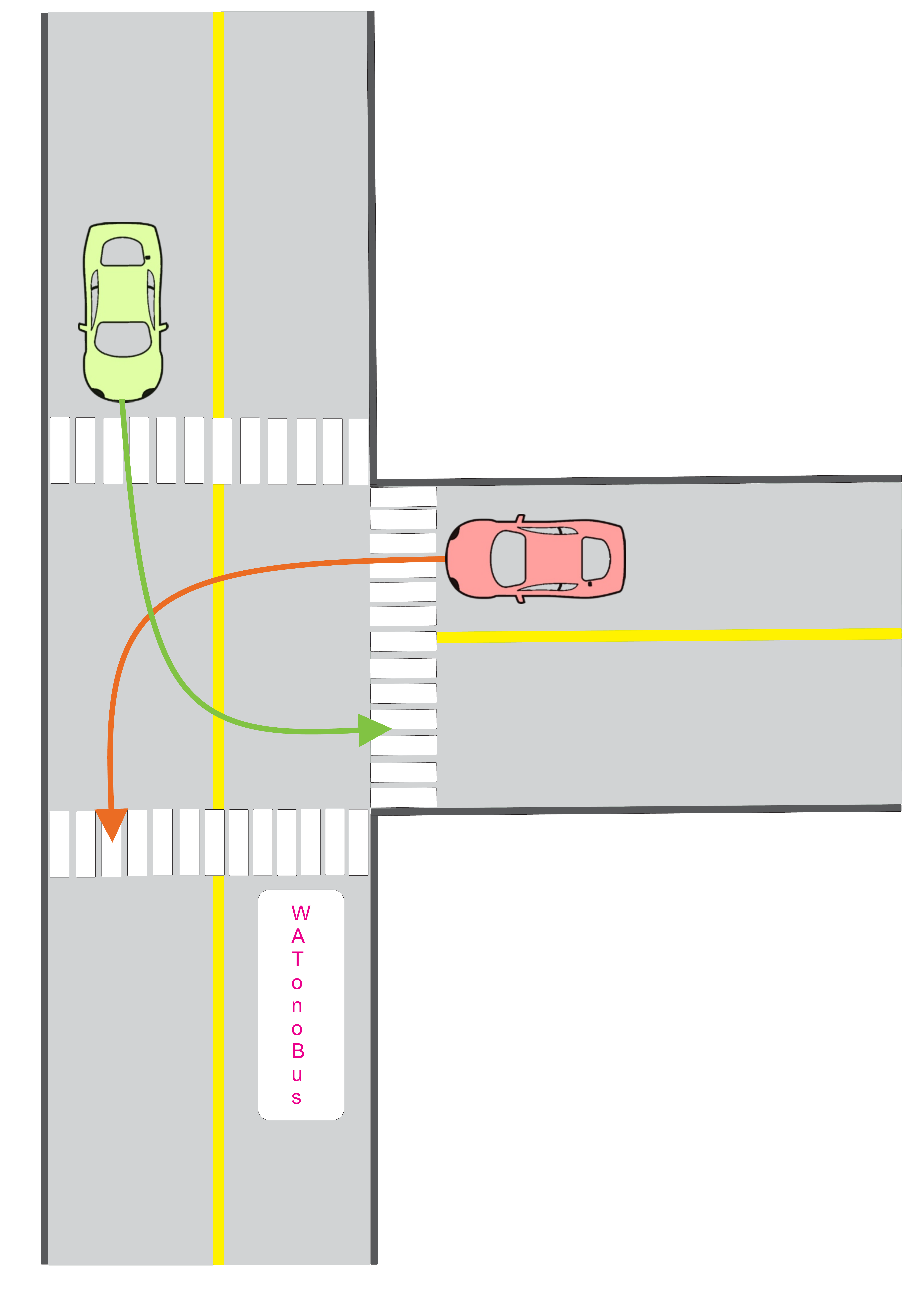}
    \caption{Vehicles}
  \end{subfigure}
  \begin{subfigure}{5cm}
    \centering\includegraphics[width=3.5cm]{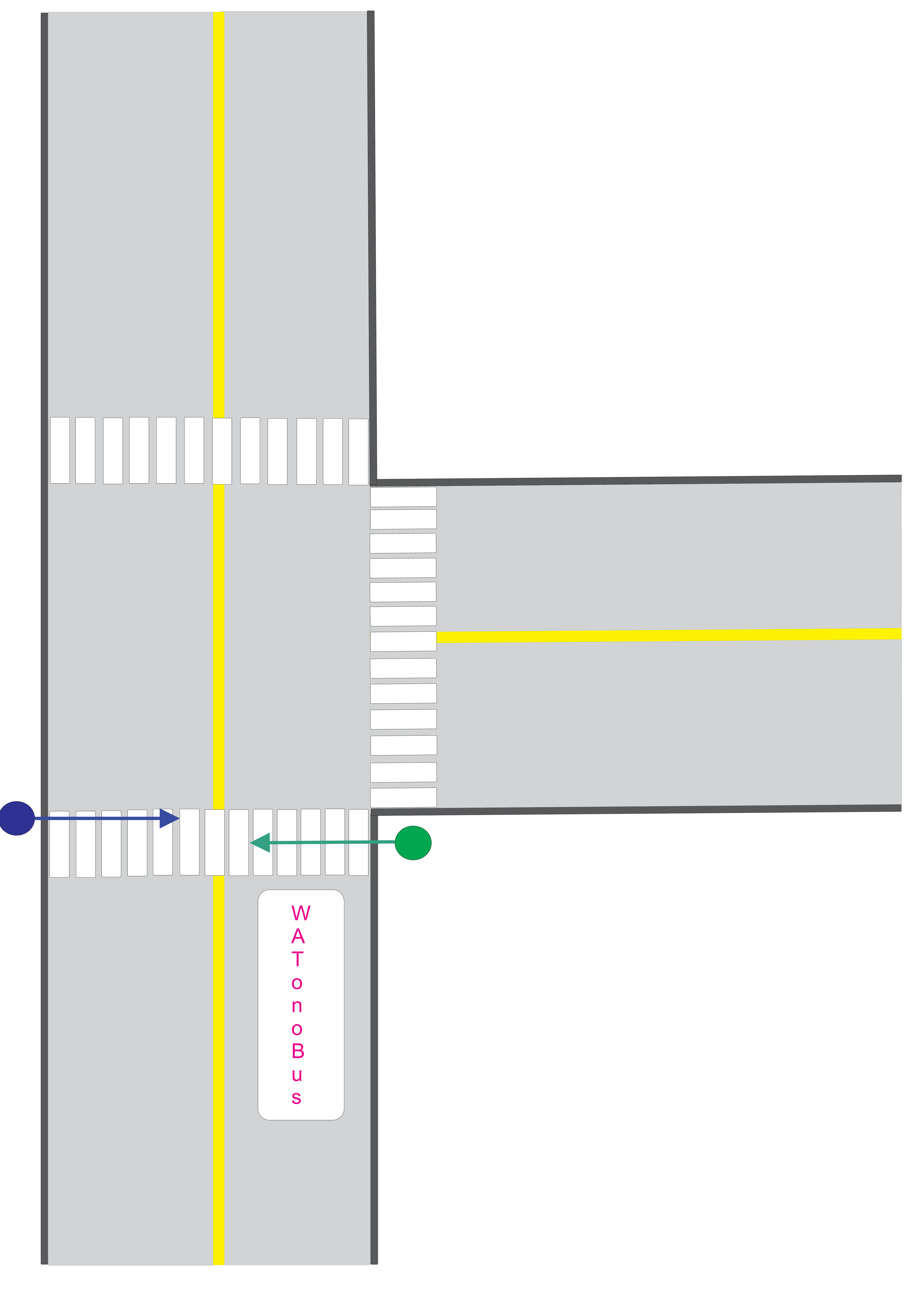}
    \caption{Pedestrians}
  \end{subfigure}
  \caption{Intersection handling.}
  \label{Figure: Intersection handling}
\end{figure}

Non-signalized intersections pose challenges due to dynamic traffic flow, right-of-way determination, and potential conflicts (See Fig. \ref{Figure: Intersection handling}). Unlike signalized intersections, they lack explicit instructions, demanding advanced perception, DM, and control capabilities for autonomous shuttle buses. Non-signalized intersections rely on right-of-way rules, which can be influenced by factors such as vehicle priority, the order of arrival. Our WATonoBus records the order of arrival for vehicle actors to decide the road priority. When comparing with pedestrian and cyclist at crossings, the shuttle consider itself lowest road priority. Moreover, non-signalized intersections introduce the possibility of conflicts arising from simultaneous movements of different road users. Conflicts can occur due to misjudgment of gaps, ambiguous driver intentions, or violations of right-of-way. The autonomous shuttle bus must be equipped with robust perception systems and prediction modules to detect and track vehicles, pedestrians, and cyclists. To reach as safe and robust behavioral planning for WATonoBus, we use an event-driven technique that reacts to potential conflicts guaranteeing the vehicle is always at safe states. 

\subsection{Perception Performance Under Snow}

\begin{figure}[t]
\centering
\includegraphics[width=0.9\linewidth]{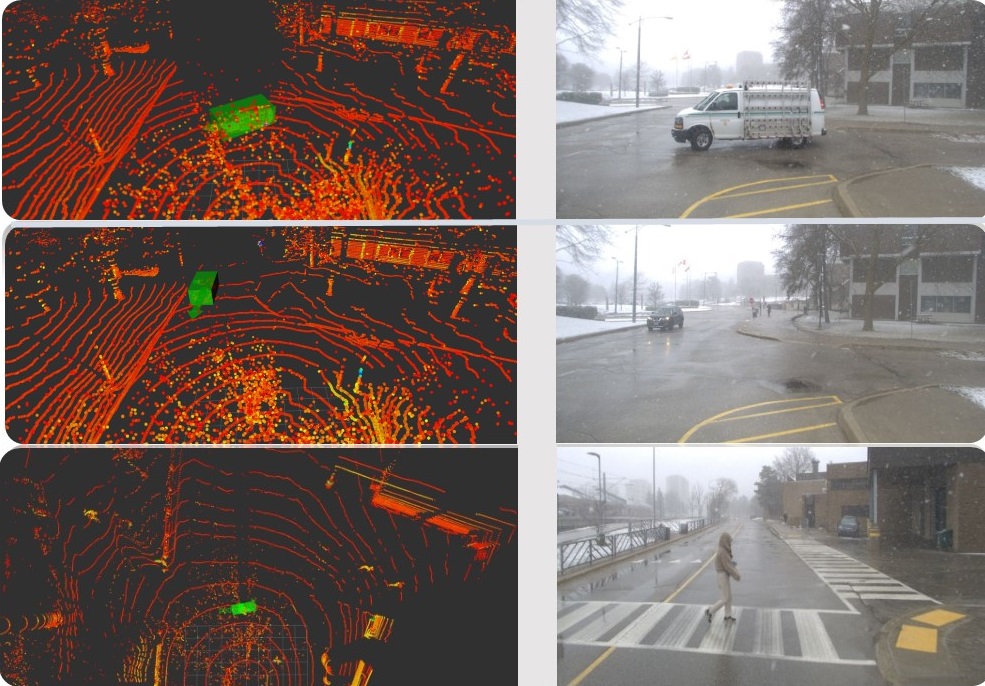}
\caption{Perception fusion results in snow.}
\label{fig:resultsPerception}
\end{figure}

Our WATonoBus perception system utilizes radar, LiDAR, and camera data fusion, particularly in snowy settings. This integrated approach successfully detects objects in three disparate yet commonplace road scenarios: a van entering the roadway, a car in the opposite lane, and a pedestrian crossing a street, as shown in Fig. \ref{fig:resultsPerception}. For detecting a van entering the road, the radar's capacity to measure distance and speed plays a crucial role, especially when fused with the camera’s high-definition imaging. The camera contributes valuable details like color and texture, allowing for a highly integrated and predictive object detection framework. This multi-sensory combination anticipates the van's potential actions, enabling dynamic adjustments for safer autonomous navigation. In the case of a car in the opposite lane, snowy conditions can introduce additional complexities like fog, glare, and reduced contrast. Here, radar's strong penetration capabilities prove indispensable for initial object detection and range estimation. Combined with the LiDAR’s noisy but detailed spatial data and the camera’s visual cues, the system doesn’t just detect the car; it precisely locates and tracks it, adding another layer of reliability and safety in treacherous weather conditions. Regarding a pedestrian on a crosswalk, snowy conditions often introduce noise into the LiDAR point clouds, making it more challenging to distinguish the shape and position of a pedestrian. In this setting, radar’s Doppler velocity data becomes invaluable. Radar's robustness to weather interference complements the high-resolution LiDAR data, filtering the noise and ensuring a reliable and comprehensive multi-modal perception of the pedestrian. This fusion mitigates the limitations posed by snowy conditions, providing an enhanced layer of safety and accuracy (\href{https://youtu.be/FePtm6bRhWA}{\textbf{video 1}} and \href{https://youtu.be/0gXi_iy9CrI}{\textbf{2}}).
% %-----------------------------%

\section{Conclusion} \label{sec:Conclusion}
%%%%%%%%%%%%%%%%%%%%%%%%%%%%%%%%%%%%%%%%%%%%%%%%%%%%%%
This paper introduces the WATonoBus project, addressing foundational research questions and design concepts with real-world experimental validation. The aim is to contribute to the continuous advancement of autonomous driving technology, providing essential insights into its feasibility, societal acceptance, and economic viability through collaborative testing and piloting. Our modular software architecture ensures all-weather functionality, specifically addressing adverse weather conditions such as rain, snow, and fog. This approach has been experimentally validated in challenging real-world scenarios. The perception module utilizes multi-modal sensor fusion for accurate object and drivable road boundary detection under adverse weather conditions, enhancing safety and reliability. The localization module demonstrates GNSS-denied capabilities, particularly crucial in challenging weather conditions. The decision-making module, supported by a dedicated safety module, ensures robust and safe autonomous operation. The intelligent bus stop pullover/merging function is tailored for shuttle bus service, adding efficiency to the shuttle's operations. Valuable insights derived from edge case learning during daily shuttle bus operations contribute to the project's continuous improvement. The construction of the distinct vehicle prototype and over a year of service experience culminate in a public unveiling, showcasing significant advancements and practical implementations in the autonomous driving domain.

%%%%%%%%%%%%%%%%%%%%%%%%%%%%%%%%%%%%%%%%%%%%%%%%%%%%%%
\section{Acknowledgement} \label{sec:Acknowledgement}
%%%%%%%%%%%%%%%%%%%%%%%%%%%%%%%%%%%%%%%%%%%%%%%%%%%%%%

The authors thank Aaron Sherratt for help with continuous testing and development as well as Jeff Graansma, Adrian Neill, and Michael Duthie. In addition, the authors would like to acknowledge the financial support of the Natural Sciences and Engineering Research Council of Canada, the Canadian Foundation of Innovation, and the Ontario Research Fund.

\bibliographystyle{IEEEtran}
{\footnotesize\bibliography{Refs}}
%\bibliography{Refs}

\end{document}